\documentclass[runningheads, a4paper]{llncs}
\usepackage[misc]{ifsym}
\usepackage{llncsdoc}
\usepackage[dvipsnames]{xcolor}
\usepackage{graphicx}
\usepackage{url}
\usepackage{multirow}
\usepackage{hyperref}

\usepackage[colorinlistoftodos,prependcaption,textsize=tiny]{todonotes}

\begin{document}

\title{Mapping Images to Psychological Similarity Spaces Using Neural Networks}
\author{Lucas Bechberger \Letter  \orcidID{0000-0002-1962-1777} \inst{1} \\
\and Elektra Kypridemou \orcidID{0000-0003-1575-9311} \inst{2}}
\institute{Institute of Cognitive Science, Osnabr\"uck University, Osnabr\"uck, Germany \email{lucas.bechberger@uni-osnabrueck.de}
\and Computational Cognition Lab, Open University of Cyprus, Nicosia, Cyprus 
\email{elektra.kypridemou@st.ouc.ac.cy}}

\authorrunning{Lucas Bechberger and Elektra Kypridemou}

\maketitle

\begin{abstract}
The cognitive framework of conceptual spaces bridges the gap between symbolic and subsymbolic AI by proposing an intermediate conceptual layer where knowledge is represented geometrically. There are two main approaches for obtaining the dimensions of this conceptual similarity space: using similarity ratings from psychological experiments and using machine learning techniques. In this paper, we propose a combination of both approaches by using psychologically derived similarity ratings to constrain the machine learning process. This way, a mapping from stimuli to conceptual spaces can be learned that is both supported by psychological data and allows generalization to unseen stimuli. The results of a first feasibility study support our proposed approach.
\end{abstract}


\section{Introduction}
\label{Introduction}

The cognitive framework of conceptual spaces \cite{Gardenfors2000,Gardenfors2014} attempts to bridge the gap between symbolic and subsymbolic AI by proposing an intermediate conceptual layer based on geometric representations.
A conceptual space is a similarity space spanned by a number of quality dimensions representing interpretable features. Convex sets in this space correspond to concepts. Abstract symbols can be grounded in perception by linking them to concepts in a conceptual space whose dimensions are based on subsymbolic representations.

The framework of conceptual spaces has been highly influential in the last 15 years within cognitive science \cite{Douven2011,Fiorini2013,Lieto2017}. It has also sparked considerable research in various subfields of artificial intelligence, ranging from robotics and computer vision \cite{Chella2003} to plausible reasoning \cite{Derrac2015}.\\

The dimensions of the conceptual similarity space are usually obtained based on either psychological experiments or machine learning methods. The first approach has a clear psychological grounding, but is only applicable to items for which such similarity ratings are available. The second approach in contrast can work on large amounts of unlabeled data, but lacks psychological validity. In this paper, we propose to combine both approaches by using psychologically derived similarity spaces as a target for machine learning algorithms. This way, a mapping from stimuli to conceptual spaces can be found that is both supported by psychological data and able to generalize to unseen stimuli. The results of a first feasibility study support our proposed approach.

The remainder of this paper is structured as follows: Section \ref{ConceptualSpaces} introduces the framework of conceptual spaces. Section \ref{PsychologicalExperiments} summarizes techniques for deriving similarity spaces from psychological experiments and Section \ref{RepresentationLearning} describes two recently proposed types of neural networks from the area of representation learning. In Section \ref{OurProposal}, we formulate our proposal of combining artificial neural networks with psychological data. Section \ref{FeasibilityStudy} presents a first feasibility study for our approach and Section \ref{Conclusions} gives an outlook on future work.

\section{Conceptual Spaces}
\label{ConceptualSpaces}

A conceptual space \cite{Gardenfors2000} is a similarity space spanned by so-called ``quality dimensions''. Each of these dimensions represents an interpretable and cognitively meaningful way in which two stimuli can be judged to be similar or different. Examples for quality dimensions include temperature, weight, time, pitch, and hue.
A domain is a set of dimensions that inherently belong together. Different perceptual modalities (like color, shape, or taste) are represented by different domains. The color domain for instance can be represented by the three dimensions hue, saturation, and brightness. Distance within a domain is measured by the Euclidean metric.
The overall conceptual space is defined as the product space of all dimensions. Distance within the overall conceptual space is measured by the Manhattan metric of the intra-domain distances.
The similarity of two points in a conceptual space is inversely related to their distance -- the closer two instances are in the conceptual space, the more similar they are considered to be.

The framework distinguishes properties like ``red'', ``round'', and ``sweet'' from full-fleshed concepts like ``apple'' or ``dog'': Properties are represented as convex sets within individual domains (e.g., color, shape, taste), whereas full-fleshed concepts span multiple domains. Reasoning within a conceptual space can be done based on geometric relationships (e.g., betweenness and similarity) and geometric operations (e.g., intersection and projection). \\

In his book \cite[Chapter 1.7, Chapter 6.5]{Gardenfors2000}, G\"ardenfors describes three ways for identifying the dimensions spanning a conceptual space: Handcrafting, multidimensional scaling, and machine learning.

Handcrafting is only possible if there is enough knowledge about the domain to be modeled. It is applicable to domains with a well known structure, for instance the temperature domain which can be described by a single dimension and measured with a single sensor. This approach is however not easily applicable to more complex domains (e.g., shapes) that are based on complex sensors (e.g., cameras). We therefore exclude it from our further considerations.

If handcrafting is not applicable, one can conduct a psychological experiment to elicit human similarity ratings for pairs of stimuli and then apply ``multidimensional scaling'' (MDS) \cite{Borg2005} which is a well-known statistical technique used in various psychological domains \cite{Jaworska2009}. MDS provides a geometric representation of the stimulus set, where geometric distances between pairs of stimuli reflect their psychological dissimilarity. We will give more detail on this approach for deriving a conceptual space in Section \ref{PsychologicalExperiments}.

The third option for obtaining the dimensions of a conceptual space is to use machine learning techniques for dimensionality reduction. G\"ardenfors argues that raw perceptual input is too rich and too unstructured for direct processing. It is thus necessary to lift the input to a more economic form of representation, which typically involves a drastic reduction in the number of dimensions. There exists a variety of dimensionality reduction algorithms in the machine learning field. Especially artificial neural networks (ANNs) are a promising candidate for implementing a multi-layered and non-linear dimensionality reduction. We will introduce two recently proposed network architectures in Section \ref{RepresentationLearning}.


\section{Deriving Conceptual Spaces from Similarity Ratings}
\label{PsychologicalExperiments}

There are several data collection methods proposed in the literature for collecting similarity judgments from human participants. The traditional technique is called \textit{``dissimilarity ratings''}. In this approach, all possible pairs from a set of stimuli are presented to participants (one pair at a time), and participants rate the dissimilarity of each pair on a continuous or categorical scale. This method is however quite inefficient, due to the large number of judgments required. 

A faster alternative is the \textit{``sorting method''}, where participants are given the whole set of stimuli and are asked to assign them into groups of similar stimuli. There are two versions of the sorting method: \textit{``constrained sorting''}, in which the number of groups is pre-defined be the experimenters, and \textit{``free sorting''}, where the number of groups is at the discretion of each participant. The resulting pairwise ratings derived from the sorting method are binary -- either the two stimuli belong to the same group or not.
Sorting methods, although faster, lack accuracy compared to dissimilarity ratings \cite{Subkoviak1976}. According to a comparative study performed on a set of 40 stimuli of sounds, dissimilarity ratings were found to be the most reliable and accurate method compared to sorting methods, lacking however in efficiency \cite{Giordano2011}.


A more recent technique named \textit{``Spatial Arrangement Method''} (SpAM) is proposed by Goldstone \cite{Goldstone1994}: Participants are asked to rate similarity through a spatial arrangement. The data set, or a part of it, is given in a random configuration and participants have to rearrange the stimuli in a two-dimensional space, such that the distance between each pair of stimuli reflects their dissimilarity.
SpAM is both efficient and user-friendly and has been found to produce high-quality MDS solutions \cite{Hout2013}. However, the SpAM method is not widely applicable, since it is constrained to visual and textual stimuli.\\

After having elicited psychological similarity ratings, MDS can be used to convert them into a geometric representation which can be interpreted as a conceptual space. MDS takes as input the pair-wise mean similarity scores for all pairs of stimuli and the desired number $n$ of dimensions. It then represents each stimulus as a point in an $n$-dimensional space and tries to arrange them in such a way that the distance of two points accurately reflects the similarity rating of the stimuli they represent.
The optimal number of dimensions is usually determined by repeatedly running the MDS algorithm with different values of $n$. The accuracy of each resulting space in representing the original similarity ratings is calculated. One typically selects a relatively small value $n$ for which the accuracy is deemed good enough and where the accuracy of using $n+1$ dimensions is not considerably better.

While providing an elegant way for deriving a conceptual space from human similarity ratings, this approach does have some disadvantages: First of all, there is no guarantee that the dimensions generated by the MDS algorithm are interpretable, so one typically needs to search for interpretable directions in the generated space. Moreover (and in our opinion more importantly), MDS is only applicable to a fixed number of input stimuli. If a new, previously unseen stimulus is perceived, MDS does not provide us with a mapping of this stimulus onto a point in the similarity space.


\section{Artificial Neural Networks for Representation Learning}
\label{RepresentationLearning}

\begin{figure}[t]
\centering
\includegraphics[width = 0.75\columnwidth]{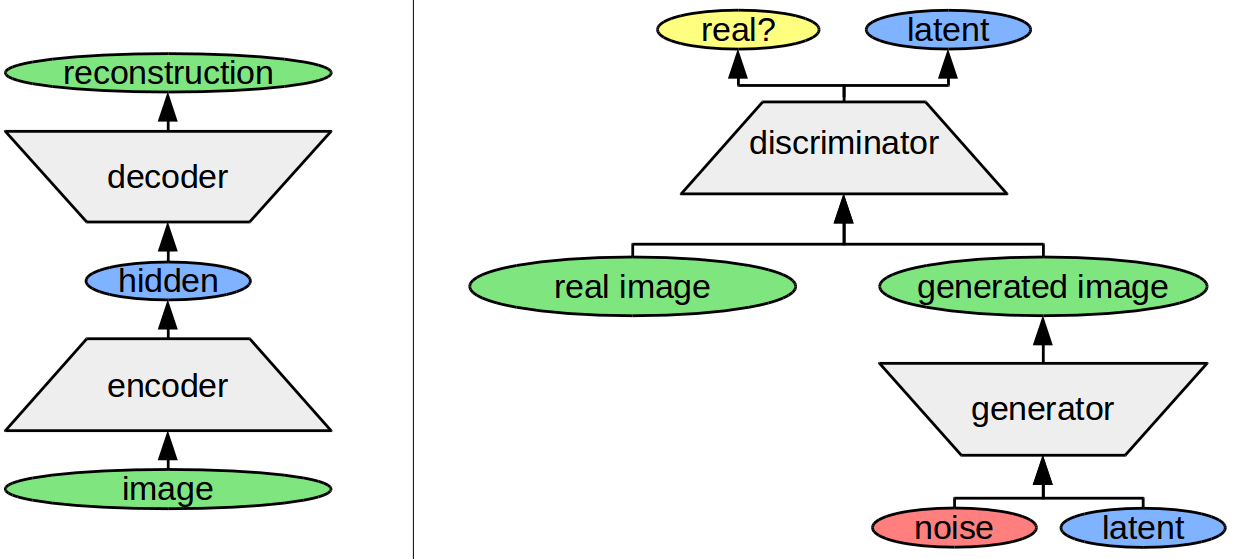} 
\caption{Left: Illustration of an autoencoder. Right: Illustration of an InfoGAN.}
\label{fig:RepresentationLearning}
\end{figure}

In recent years, there has been substantial work on learning compressed representations of a given feature space with neural networks \cite{Bengio2013}. We focus our discussion on two recently proposed approaches that have shown the potential to extract interpretable dimensions, namely $\beta$-VAE \cite{Higgins2017} and InfoGAN \cite{Chen2016}.\\

The left part of Figure \ref{fig:RepresentationLearning} shows the typical structure of an autoencoder. The overall network consists of two parts, an encoder and a decoder. The encoder takes an input image and compresses it into a low-dimensional ``hidden'' representation. This hidden representation is passed on to the decoder, which tries to reconstruct the original image from it.

While the general autoencoder structure has existed for decades, Kingma and Welling \cite{Kingma2013} have recently proposed variational autoencoders (VAE) that use Gaussian distributions as their hidden representation. The encoder predicts two numbers for each entry of the hidden representation, which are then interpreted as the mean and the variance of a Gaussian distribution. The input to the decoder is sampled from this distribution. The loss term that the network tries to minimize consists of two parts: The reconstruction error measures the difference between the original input image and its reconstruction, and the regularization term measures the difference between a multivariate Gaussian distribution with a diagonal covariance matrix and the actual distribution of the hidden layer. The experimental results of Kingma and Welling showed that this architecture is able to learn very good reconstructions of the original input images.

Recently, Higgins et al \cite{Higgins2017} have modified this framework by giving a larger weight $\beta$ to the regularization term than to than the reconstruction error. Their experiments showed that the resulting $\beta$-VAE network is able to extract meaningful dimensions from unlabeled data sets because the network has a stronger incentive to learn uncorrelated hidden dimensions. However, due to the smaller emphasis on the reconstruction error, this improved interpretability of the extracted dimensions is gained by sacrificing reconstruction accuracy.\\

The InfoGAN network \cite{Chen2016} (depicted in the right part of Figure \ref{fig:RepresentationLearning}) is based on the framework of generative adversarial networks (GAN) and consists of two networks, the generator and the discriminator. The generator is fed with two low-dimensional vectors of noise values. Its task is to create high-dimensional data vectors that have a similar distribution as real data vectors taken from an unlabeled training set. The discriminator receives a data vector that was either created by the generator or taken from the training set (e.g., images). Its task is to distinguish real inputs from generated inputs and to reconstruct one of the noise vectors (the so called ``latent'' vector). The overall architecture can be interpreted as a two-player game: The generator tries to fool the discriminator by creating realistic images and the discriminator tries to avoid being fooled by the generator. Chen et al. \cite{Chen2016} showed that after training an InfoGAN, the latent variables tend to have an interpretable meaning. For instance, in an experiment using the MNIST data set, the latent variables corresponded to type of digit, digit rotation and stroke thickness. In practice, however, InfoGAN and other GAN variants are relatively difficult to train \cite{Arjovsky2017}.\\

Both of these recent approaches have shown promising early results in extracting meaningful dimensions from unlabeled data. They can both generalize to unseen inputs and generate example images from a given hidden/latent representation. However, choosing the correct size of the hidden/latent representation is crucial and typically requires either good prior knowledge of the domain or extensive manual optimization. Neural networks in general require large amounts of training data, but as both presented approaches only need unlabeled data, this is not critical in practice. Finally, while the extracted dimensions might be useful from an AI perspective, they cannot claim any psychological validity.

\section{Our Proposal}
\label{OurProposal}

As we have seen in Sections \ref{PsychologicalExperiments} and \ref{RepresentationLearning}, both ``traditional'' approaches for obtaining the dimensions of a conceptual space have their individual strengths and weaknesses. We are not aware of an approach that can claim both to be psychologically valid and to generalize well to unseen stimuli. By combining MDS with neural networks, one can potentially obtain such an approach. We will focus our subsequent discussion on stimuli in the form of images.

\subsection{Proposed Procedure}
\label{OurProposal:Procedure}

After having determined the domain of interest (e.g., the domain of shapes), one first needs to acquire a data set of stimuli from this domain (e.g., ShapeNet \cite{Chang2015}). This data set should cover a wide variety of stimuli and it should be large enough for applying machine learning algorithms. Using the whole data set with thousands of stimuli in a psychological experiment is unfeasible in practice. Therefore, a relatively small, but still sufficiently representative subset of these stimuli needs to be selected for the elicitation of human similarity ratings. 

This subset of stimuli is then used in a psychological study where similarity judgments by humans are obtained, using one of the techniques described in Section \ref{PsychologicalExperiments}. The choice of the data collection method usually depends on the type of stimuli, the size of the data set and the aims of the study. Choosing the right method is crucial for the quality of the resulting space \cite{Subkoviak1976}.

One can now apply MDS to the similarity judgments from the psychological experiment in order to extract a spatial representation of the underlying domain. As stated in Section \ref{PsychologicalExperiments}, one needs to select the optimal number of dimensions -- either based on prior knowledge or by manually optimizing the trade-off between high representational accuracy and a low number of dimensions. From a machine learning perspective, the stimulus-point mappings obtained from MDS can be interpreted as labeled training instances: The stimulus is identified with the input vector and the point in the MDS space is regarded as its label. 

Now an ANN can be trained to map from the stimuli (e.g., images) to the points in the MDS space. As the number of stimulus-point pairs is quite low for a machine learning problem, one needs to introduce an additional training objective (e.g., minimizing the reconstruction error on the full data set) to avoid overfitting and to ensure that the network is able to generalize to unseen inputs.

If training the ANN is successful, then it has learned a mapping from stimuli to points in the conceptual space. While MDS only maps from a fixed set of stimuli to points in the conceptual space, the ANN is expected to generalize this mapping to previously unseen stimuli. Moreover, as psychologically derived data was used to train the network, it can claim at least some psychological validity.

\subsection{Possible Network Architectures}
\label{OurProposal:NetworkArchitectures}

\begin{figure}[t]
\centering
\includegraphics[width = 1.0\columnwidth]{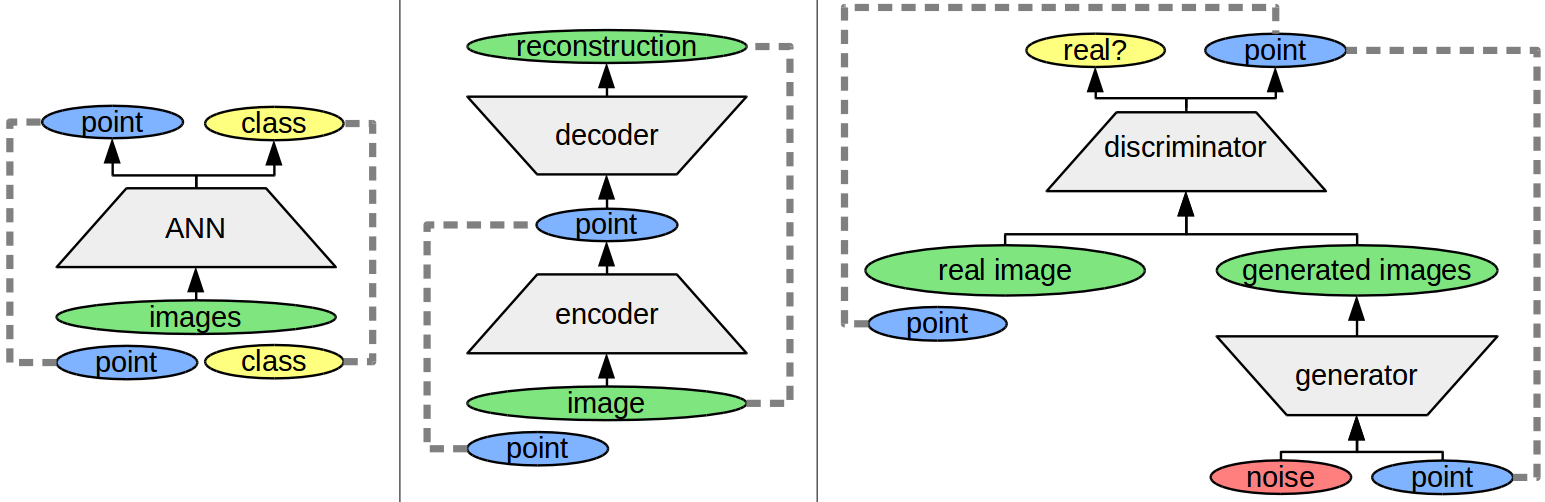} 
\caption{Three potential network architectures for our proposed learning problem. Constraints are shown by the dashed grey lines. Left: A standard feed-forward network trained also on a secondary task like classification. Middle: A (variational) autoencoder with a constrained bottleneck layer. Right: An InfoGAN network with an additional reconstruction constraint for labeled images.}
\label{fig:NetworkArchitectures}
\end{figure}

Figure \ref{fig:NetworkArchitectures} illustrates three potential network architectures for image-point mappings, where the dashed grey lines illustrate the reconstruction constraints used in the training procedure. In all cases, the respective network is trained on a secondary task for all images from the full data set and on the mapping task only for the images that were used in the psychological study. Using a secondary task with additional training data constrains the network's weights and can be seen as a form of regularization: These additional constraints are expected to counteract overfitting tendencies, i.e., tendencies to memorize all given mapping examples without being able to generalize.

In the left part of Figure \ref{fig:NetworkArchitectures}, a standard feed-forward network is used to learn a mapping from images to points, supported by a secondary task of predicting the correct classes. This approach is however only applicable if a large data set with class labels is available.

The middle part of Figure \ref{fig:NetworkArchitectures} shows the network structure of an autoencoder: The network is trained to minimize the difference between the input images and their reconstruction, while being encouraged to use the MDS space in its bottleneck layer. As the computation of the reconstruction error does not need any class labels, this is applicable also to unlabeled data sets. Of course, instead of using a regular autoencoder, one can also employ VAE or $\beta$-VAE. The autoencoder structure has the additional advantage that one can use the decoder network to generate an image based on a point in the conceptual space.

The right part of Figure \ref{fig:NetworkArchitectures} shows an extended version of the InfoGAN framework, where the discriminator network is also constrained to correctly extract the MDS points from the images. This is in some sense reminiscent of the AC-GAN network \cite{Odena2017} which however predicts categorical class labels instead of continuous points in a semantic space. Also this network architecture is capable of generating images based on a given point in the conceptual space.\\

Instead of learning a direct mapping from images to points in a conceptual space, one can also train a network to predict the similarity ratings for pairs of images. In this case, the neural network receives two images as input and predicts their similarity. The input dimensionality is twice as large in this case, but there are also more training examples available. Again, an additional training objective might be needed in order to prevent overfitting. In principle, the three approaches discussed above can be adapted also to this learning problem. One can use such a network to indirectly map an image onto a point in the conceptual space as follows: The network predicts the similarity of the given image to a fixed set of ``anchor images'' for which a mapping into the conceptual space is known. One can then use the points representing these anchor images together with the predicted similarity ratings to triangulate the point representing the new image.

\subsection{Discussion}
\label{OurProposal:Discussion}

What are the advantages of our proposed approach? As stated before, we expect to obtain a neural network that learns a mapping from input images to a psychological similarity space. We thus get a mapping that is both psychologically grounded and that generalizes to unseen inputs. Even if the mapping performance of the neural network is not perfect, a rough guess for the point representing a previously unseen stimulus might still be quite useful.\\

In general, one can distinguish two types of similarity: While perceptual similarity is exclusively based on the immediate perceptual features of the stimuli, conceptual similarity involves additional sources of knowledge, e.g., the expected use of a depicted object or the typical context in which it can be found.

Neural networks that are trained on a task such as reconstruction are likely to yield a hidden representation which encodes perceptual similarity. The human similarity ratings retrieved in psychological experiments might however also include conceptual similarity, and thus implicit features such as the perceived object's softness. If a neural network is not only trained to reconstruct images, but also to predict their positions in a psychological similarity space, it is incited to use also such implicit features in its internal representations. In some sense, these psychological constraints could thus help the network to ``see behind the image'' and to represent not only perceptual, but also conceptual similarities.\\

If the approach described above yields promising and useful results, one could use the observation that $\beta$-VAE and InfoGAN tend to discover meaningful dimensions in order to devise a new MDS algorithm based on these networks. This algorithm would train a standard $\beta$-VAE or InfoGAN network while ensuring through an additional term in the loss function that the hidden space extracted by the network accurately reflects the psychological similarity ratings. This overall algorithm would thus result in a spatial representation of psychological similarities which generalizes to unseen images, uses interpretable dimensions, and can be used to generate new images based on points in the conceptual space.\\


\section{Feasibility Study}
\label{FeasibilityStudy}

In order to validate whether our proposed approach is worth pursuing and in order to support our claims with first empirical results, we conducted a feasibility study based on a simple setup: We used an existing data set of similarity ratings for images of novel objects and we trained a linear regression on the hidden activations of a pre-trained image classification network.\footnote{Code for reproducing this study can be found online at \url{https://github.com/lbechberger/LearningPsychologicalSpaces/} \cite{Bechberger2018GitHubPsy}.}

\subsection{Data Set}
\label{FeasibilityStudy:DataSet}

For our feasibility study, we used existing similarity ratings reported for the Novel Object and Unusual Name (NOUN) dataset \cite{Horst2016}, a set of 64 images of 3-D objects that are designed to be novel but also look naturalistic. Adopting the SpAM approach \cite{Goldstone1994}, participants were presented with 13 trials of 20 objects, assigned randomly in a way such that all pairwise comparisons among the 64 images were evaluated. On each trial, stimuli were presented to participants on a 4x5 matrix arrangement and participants re-arranged the images such that the distance between each pair of images in the final configuration reflected their dissimilarity. Based on the coordinates of each image in the final configurations, an individual similarity matrix was calculated for each participant. The similarity matrices of all participants were then averaged into a global similarity matrix by using SPSS and the PROXSCAL algorithm.

\begin{figure}[t]
\centering
\includegraphics[width = 1.0\columnwidth]{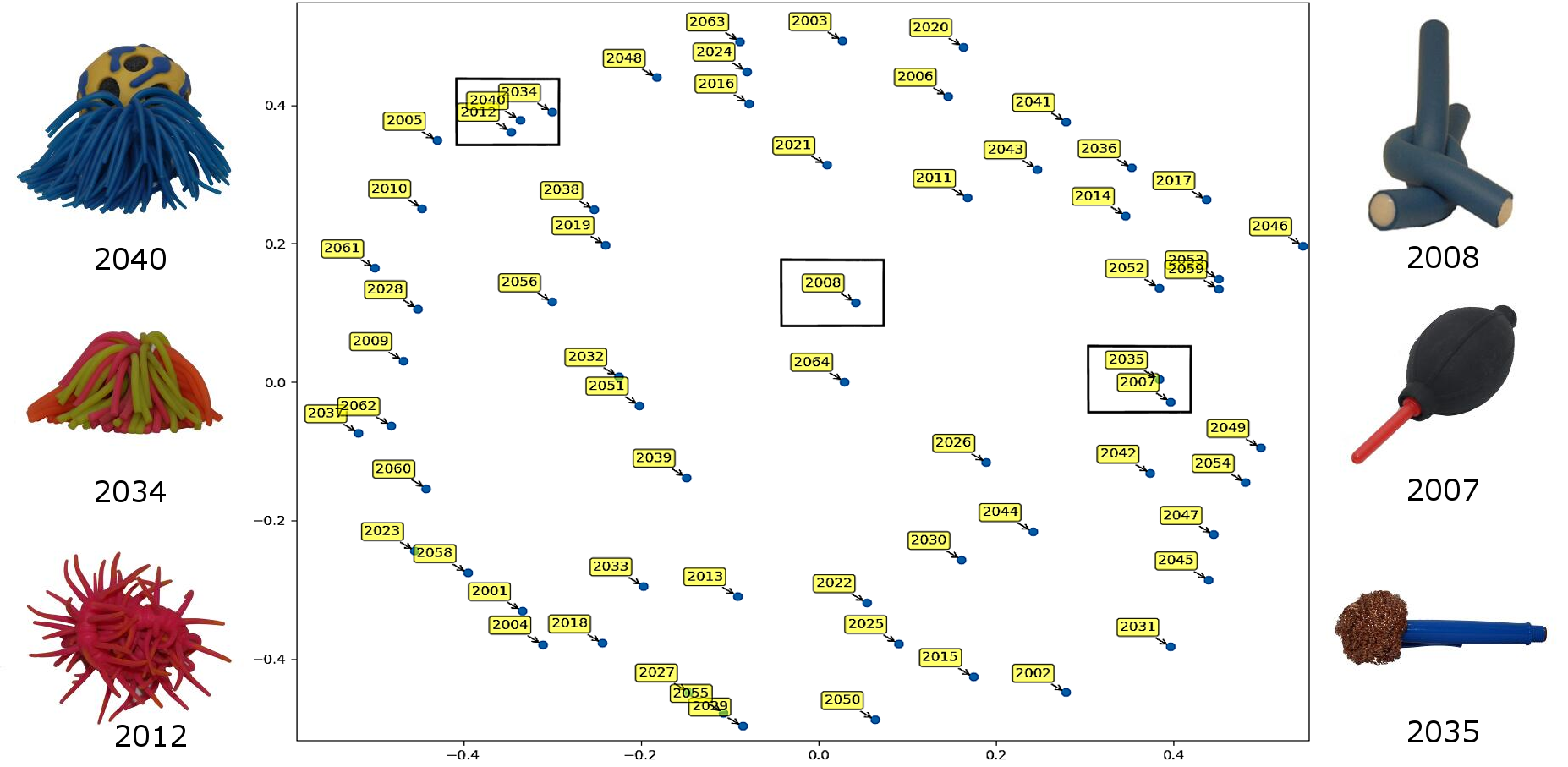} 
\caption{Two-dimensional MDS space derived from the NOUN data set. Examples of the images are given for the areas displayed in the rectangles.}
\label{fig:mapping}
\end{figure}

Using the similarity matrix from the study of Horst and Hout \cite{Horst2016}, we performed a metric MDS by running the SMACOF algorithm for four times and keeping the best results, with a maximum of 300 iterations per run, using the precomputed dissimilarity measure. In the work by Horst and Hout, a four-dimensional space was found to be the best trade-off between a low-dimensional space and a good reflection of the original similarity scores. We ran two additional variations of the MDS, resulting in similarity spaces with two, four, and eight dimensions, respectively. Figure \ref{fig:mapping} illustrates the position of some example images in the two-dimensional space.\\

As the number of images is quite low for a machine learning problem, we augmented the data set by applying horizontal flips, random crops, a Gaussian blur, manipulation of the image's contrast and brightness, additive Gaussian noise, affine transformations, and salt and pepper noise. For each of the original 64 images, we created 1,000 augmented versions, resulting in a data set of 64,000 images in total. We assigned the MDS point of the original image to each of the 1,000 augmented versions.

\subsection{Experimental Setup}
\label{FeasibilityStudy:ExperimentalSetup}

Due to the still limited variability of our training examples, we decided to use a pre-trained network instead of training a new network from scratch. We used the Inception-v3 network \cite{Szegedy2016}, which has been trained on ImageNet \cite{Deng2009} and which has achieved a state of the art top-5 error rate of 3.46\% when classifying images into one of 1000 classes.
We used the activations of the second-to-last layer as a 2048-dimensional feature vector and trained a linear regression to map from this feature vector to the points in the MDS space. We assume that this feature vector, although derived from a different task, still provides enough information.

Our network architecture is a special variant of the feed-forward network proposed in Section \ref{OurProposal:NetworkArchitectures}: Instead of training both the mapping and the classification task simultaneously, we use an already pre-trained network, keep its weights fixed, and augment it by an additional output layer which is trained separately.

We have implemented four baselines against which we compare our system:
\vspace{-\topsep}
\begin{itemize}
    \item \textbf{Zero}: Always predict the origin, i.e., $(0, 0, \dots, 0)$.
    \item \textbf{Mean}: Always predict the mean of all MDS points from the training set.
    \item \textbf{Distribution}: Draw a random sample from a Gaussian distribution which was estimated from the MDS points in the training set.
    \item \textbf{Random Draw}: Use a randomly selected MDS point from the training set as prediction.
\end{itemize}
\vspace{-\topsep}

We expect that our system is able to outperform all of these baselines. We would however also like to investigate whether learning a mapping into an MDS space is easier than learning a mapping into an arbitrary space of the same dimensionality.
Therefore, we trained and evaluated two versions of our system: One of them was trained on the mappings from images to points derived by the MDS. For the other version, we used the same MDS points, but we shuffled the assignment of images to points. We ensured that all augmented images created based on the same original image were still mapped onto the same MDS point. With this shuffling procedure, we aimed to destroy any semantic structure inherent in the original space. We expect that the regression works better for the original mapping than for the shuffled mapping, as it contains more semantic content and should therefore be easier to learn. By using a shuffled mapping rather than a mapping to randomly chosen points in the space, we ensure that the distribution of the target points remains constant. This is necessary in order to make a meaningful comparison between the two regression configurations.\\

In order to evaluate both our system and the baselines, we used the root mean squared error (RMSE), which is a standard metric for regression problems. The RMSE metric has the same scale as the target space and can thus provide an intuition about the system's performance if interpreted as the average distance between the prediction and the ground truth.
Due to the small amount of original images, we performed an image-based leave-one-out evaluation: All augmented images generated from one of the original images were used as test set, while all other images were to train the linear regression. This was repeated for each of the original images and the training and test errors were averaged across all of these runs. As all augmented images based on the same original image can be expected to still be quite similar to each other, this procedure ensures that the system cannot simply ``memorize'' all 64 images.

\subsection{Results}
\label{FeasibilityStudy:Results}

\begin{table}[t]
  \centering
  \begin{tabular}{|l||c|c||c|c||c|c|}
    \hline
    \multirow{2}{*}{\textbf{Configuration}}      & \multicolumn{2}{|c||}{\textbf{2D}} & \multicolumn{2}{|c||}{\textbf{4D}} & \multicolumn{2}{|c|}{\textbf{8D}} \\ 
                	                    & Training      & Testing      & Training      & Testing      & Training      & Testing \\ \hline \hline
    \textbf{Zero baseline}	                    & 0.4408        & 0.4408    & 0.4596        & 0.4596    & 0.4595        & 0.4595 \\ \hline
    \textbf{Mean baseline}	                    & 0.4408        & 0.4478    & 0.4595        & 0.4669    & 0.4594        & 0.4668 \\ \hline
    \textbf{Distribution baseline} 	            & 0.6287        & 0.6189    & 0.6540        & 0.6449    & 0.6520        & 0.6404 \\ \hline
    \textbf{Random draw baseline} 	            & 0.6233        & 0.6283    & 0.6499        & 0.6554    & 0.6498        & 0.6551 \\ \hline \hline
    \textbf{Regression (shuffled)}	& 0.0673        & 0.3327    & 0.0501        & 0.2472    & 0.0366        & 0.1893 \\ \hline
    \textbf{Regression (correct)}	    & 0.0534        & 0.2274    & 0.0412        & 0.1779    & 0.0295        & 0.1287 \\ \hline
  \end{tabular}\\[2ex]
\caption{Mean RMSE on both the training and the testing data for the baselines and the regression, targeting a two-, four- and eight-dimensional space, respectively. Each RMSE was averaged over ten independent runs.}
\label{tab:ExperimentalResults}
\end{table}

Table \ref{tab:ExperimentalResults} shows the results obtained in our experiment for the three MDS spaces with two, four, and eight dimensions, respectively. Among the different baselines, the ``zero baseline'' consistently has the best performance on the test data, followed closely by the ``mean baseline''. The two other baselines perform considerably worse. Notably, the regression always has a much lower RMSE than any of the baselines, indicating that the system is indeed capable of learning a mapping from images to points in a psychological space. As expected, the RMSE for the correct mapping is always considerably lower than for the shuffled mapping. However, the regression always performs significantly better on the training data than on the test data, which is a clear sign for overfitting.

\begin{figure}[t]
\centering
\includegraphics[width = 0.9\columnwidth]{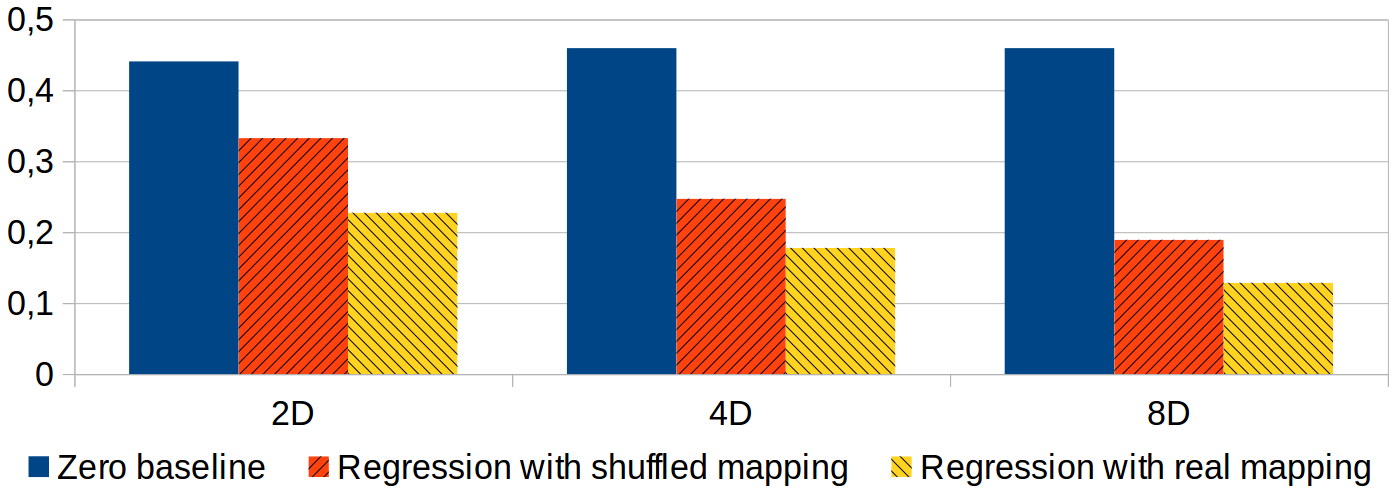} 
\caption{Graphical illustration of the RMSE of the test data as reported in Table \ref{tab:ExperimentalResults}.}
\label{fig:Results}
\end{figure}

Figure \ref{fig:Results} illustrates the RMSE results during testing for the three different similarity spaces. As one can see, the performance of the best baseline is quite independent from the dimensionality of the similarity space. Although the regression problem in theory becomes harder with an increasing number of dimensions (when predicting more coefficients of the output vector, one can make more mistakes), both regression approaches perform better in higher-dimensional spaces. 

\subsection{Discussion}
\label{FeasibilityStudy:Discussion}

The results presented in Section \ref{FeasibilityStudy:Results} confirm our hypotheses: Our system is able to perform better than the baselines, which shows that a mapping from images to MDS points can be learned. Moreover, the observed performance difference on the test set between learning the correct mapping and learning the shuffled mapping indicates that the semantic structure of a similarity space facilitates learning a generalizable mapping. The small difference on the training data indicates that both learning problems have a similar difficulty.

The regression also seems to benefit from a higher-dimensional MDS space. As a higher-dimensional space derived with MDS can more accurately reflect the original similarity ratings, this might be interpreted as the system learning finer nuances of similarity if given the chance to do so. As all baselines have a similar performance on all MDS spaces, the distribution of the points in the space seems to have similar characteristics independent of the number of dimensions.
We assume that also the regression on the shuffled mapping can benefit from a higher number of dimensions as the shuffling procedure may not be able to completely destroy the structure of the similarity space -- even after shuffling, the mapping might still partially reflect the similarity ratings.\\

These overall results are in line with our expectations and indicate that our overall approach is sound and promising.
The achieved performance is still far from perfect, probably because of the observed overfitting tendencies. However, the goal of this early feasibility study was not to aim for optimal performance, but to evaluate the principle idea of our approach. 
We suspect that the relatively large amount of overfitting is caused by the small set of original images and the large size of the extracted feature vector. Although we applied a variety of augmentation techniques to the original images, the resulting images might still be too similar to each other. Moreover, as the feature vector contains 2048 entries, the linear regression has to estimate a large number of parameters.
Furthermore, as the network was pre-trained on a different data set with different characteristics, we cannot expect a perfect generalization.

Finally, one could argue that even humans might fail to achieve an RMSE of approximately zero, as the given problem is inherently difficult, due to the nature of the data set used. The MDS space was created based on the average similarity ratings obtained in Horst and Hout's \cite{Horst2016} study using a set of novel unknown images, for which the similarity judgments might differ a lot among individuals. 

\section{Conclusion and Future Work}
\label{Conclusions}

We have argued that the combination of neural networks with psychological similarity judgments offers a promising way of extracting a conceptual space from data. This can help to make the framework of conceptual spaces more viable for artificial intelligence: Our proposed approach can potentially provide a principled way of mapping sensory input to conceptual spaces, while still maintaining some psychological validity. The results of our feasibility study are encouraging and show that our approach is also feasible in practice.

In future work, we will implement and evaluate our proposal in more depth by exploring the remaining proposed network structures. We will conduct a psychological study on a subset of ImageNet \cite{Deng2009} to investigate whether we can achieve a better mapping performance if we use images from the same domain.
The data used in our current study has the shortcoming of not separating different domains like color, shape, and size (as proposed by G\"ardenfors), but of treating them as a single space. In order to evaluate whether this difference impacts the performance of our proposal, we will apply our approach to a single domain (namely, shapes).
Finally, we will also explore additional ways of evaluating the mapping performance (e.g., by comparing to the original similarity ratings).

\bibliographystyle{splncs03}
\bibliography{bibliography.bib}
\end{document}